\documentclass[conference,a4paper,11pt]{IEEEtran}

\usepackage[utf8]{inputenc}
\usepackage[T1]{fontenc}
\usepackage{graphicx}
\usepackage{amsmath}
\usepackage{booktabs}
\usepackage{cite}
\usepackage{hyperref}
\usepackage{xcolor}
\usepackage{etoolbox}

\newcommand{\rev}[1]{{\color{black}#1}}

\makeatletter
\let\rev@oldbibitem\bibitem
\renewcommand{\bibitem}[1]{\normalcolor\rev@oldbibitem{#1}%
  \ifstrequal{#1}{balfaqih2026}{\color{blue}}{}}
\makeatother

\begin{document}

\title{A Reliability Study of Explainable AI\\
for Freight Mode Classification}

\author{
\IEEEauthorblockN{Ali Alawi Balfaqih}
\IEEEauthorblockA{AI Verification and Validation (AI-VNV) Lab\\
King Fahd University of Petroleum \& Minerals (KFUPM)\\
Dhahran, Saudi Arabia\\
Email: guest60@kfupm.edu.sa}
\and
\IEEEauthorblockN{Mansur M. Arief}
\IEEEauthorblockA{Department of Industrial and Systems
Engineering, \\
IRC for Smart Mobility and Logistics (IRC-SML), \\ and AI-VNV Lab, KFUPM, 
Saudi Arabia\\
Email: mansur.arief@kfupm.edu.sa}
}

\maketitle

\begin{abstract}
Machine learning models are increasingly used to support freight
transportation decisions, but their predictions are rarely accompanied by
evidence that the associated explanations can be trusted. In this study we
investigate explainable machine learning for classifying shipments as
Air/Air Charter or Truck/Ocean using the PEPFAR Supply Chain Management
System Delivery History dataset, with 5,964 shipment records and 47 input
features after cleaning, where freight cost and pack price are excluded
as near-proxies of the transportation mode. We compare three Multi-Layer
Perceptron (MLP) architectures, Logistic Regression, Random Forest, and
gradient boosting with imbalance-aware metrics and paired significance
tests across one hundred random replications. We then evaluate the explanations
themselves, namely the stability of SHAP feature rankings across
retraining and across models, the faithfulness of SHAP and LIME
attributions under a feature-deletion test, and the geographic structure
of the destination effects, which we map for all 38 destination countries. The results suggest that the
tree ensembles significantly outperform the neural and linear models
(86.8\% against 83.3--85.0\% mean testing accuracy) while their leading
features remain the most stable across retraining. Removing the five
features ranked most important by SHAP or LIME changes the predicted
probabilities roughly four times as much as removing randomly chosen
features. The destination effects form a coherent geographic pattern
that simple port access does not explain. However, models with similar
accuracy disagree
substantially on which features matter most, and thus explanation
stability and faithfulness should be assessed before such explanations
are used to support freight transport decisions.
\end{abstract}


\section{Introduction}

Freight transportation is a fundamental component of modern supply chain
management, enabling the movement of goods between suppliers,
manufacturers, distributors, and customers. Selecting an appropriate
transportation mode is a complex decision influenced by multiple shipment
characteristics, including weight, cost, delivery time, distance, value,
and product attributes, and it interacts with other logistics decisions
such as shipment planning and storage capacity under uncertainty
\cite{pujawan2015}. As supply chains become
increasingly data-driven, machine learning (ML) techniques have been widely
adopted to support transportation-related decision making because they can
capture complex and nonlinear relationships that are difficult to represent
with traditional methods \cite{olan2025,xu2024}.

While predictive performance is important, it is equally critical that
practitioners understand why a model produces a particular prediction
before relying on it in real operations, in the same way that black-box
systems in safety-critical settings require rigorous evaluation before
deployment \cite{arief2021}. Explainable Artificial Intelligence (XAI)
has therefore emerged as a research area that aims to improve model
transparency for complex models
\cite{schwalbe2024,olan2025,ribeiro2016}. Among the various XAI
techniques, SHapley Additive exPlanations (SHAP) \cite{lundberg2017}
assigns each feature a game-theoretic contribution value for an
individual prediction, while Local Interpretable Model-agnostic
Explanations (LIME) \cite{ribeiro2016} approximates the model with a
simple interpretable surrogate in the neighborhood of a selected
observation.

There is a growing body of literature applying these techniques in
logistics and transportation. Xu et al.~\cite{xu2024} showed that
interpretable ML methods can reveal meaningful relationships between
shipment characteristics and freight mode selection, Olan et
al.~\cite{olan2025} discussed how explainable AI can improve transparency
in supply chain decision support, and Ogorodnyk et
al.~\cite{ogorodnyk2025} employed Shapley values together with LIME to
explain reinforcement learning decisions in delivery route optimization.
These studies demonstrate the practical value of explanations, but they
generally report the explanations as produced, without examining their
reliability.

However, there are at least two reasons to be cautious before acting on
such explanations. \textit{First}, many different models can fit the same
data almost equally well, a phenomenon known as the Rashomon effect
\cite{breiman2001}, and post-hoc methods applied to these
equally accurate models may attribute the predictions to different
features, a disagreement problem documented among practitioners
\cite{krishna2022} and confirmed in comparative studies of SHAP and LIME
\cite{salih2025}. \textit{Second}, an explanation is only useful if it is
faithful, i.e., if the features it highlights are the features the model
actually relies on, and faithfulness is rarely quantified in logistics
applications even though established evaluation procedures exist
\cite{samek2017}. In addition, earlier freight mode classification
studies relied mainly on
accuracy, which can be misleading under class imbalance \cite{saito2015}.

Our research is an attempt to fill the above gaps by treating the
explanations, and not only the predictions, as objects of evaluation. The
contribution of this paper is fourfold. First, we extend the model
comparison for shipment mode classification with tree-ensemble baselines,
imbalance-aware metrics, and paired significance tests across one hundred
random replications. Second, we quantify the stability of global SHAP feature
rankings across retraining replications and across models, which provides
direct evidence on the Rashomon effect in a freight transportation
setting. Third, we evaluate the faithfulness of SHAP and LIME
attributions with a feature-deletion test against a random-ranking
baseline. Fourth, we encode every destination country individually and
map its contribution to the mode prediction on an equal-area projection,
which connects the explanations to corridor-level logistics conditions.
All experiments use self-contained implementations of Kernel SHAP and
LIME that we validate against exact closed-form SHAP values for the
linear model.

\section{Data and Classification Models}

\subsection{Dataset and Prediction Target}

The analysis uses the public PEPFAR Supply Chain Management System (SCMS)
Delivery History dataset, published by the United States Agency for
International Development (USAID) \cite{usaidscms}, which contains
10,324 international shipment records of antiretroviral medicines, rapid
test kits, and other health commodities. Records with missing shipment-mode values or unusable entries in the
required numerical variables were removed, leaving 5,964 observations for development and evaluation.

The original \textit{Shipment Mode} variable includes Air, Air Charter,
Truck, and Ocean. We transformed the prediction task into a binary
classification problem in which Air and Air Charter records form class 1
and Truck and Ocean records form class 0. Approximately 75.8\% of the
cleaned records belong to the Air class, so the class distribution is
clearly imbalanced.

\subsection{Input Features and Preprocessing}

Four numerical variables serve as model inputs, namely \textit{Line Item
Quantity}, \textit{Line Item Value}, \textit{Unit Price}, and
\textit{Weight (Kilograms)}, together with two categorical variables,
\textit{Country} and \textit{Product Group}. All 38 destination
countries in the cleaned data are one-hot encoded as individual
indicators, \rev{together with the five product groups (ARV, HRDT, ANTM,
ACT, and MRDT),} and the final input data contain 47 features\rev{, of
which 43 are binary indicators}. Note that \textit{Freight Cost
(USD)} and \textit{Pack Price} are excluded from all trained models,
because freight cost directly reflects the cost structure of the chosen
mode and pack price acted as a dominant value-related signal of air
shipment in our preliminary experiments, so removing both forces the
models to rely on shipment characteristics instead of near-proxies of
the label.

The preprocessing and the classifiers are implemented with Scikit-learn
\cite{pedregosa2011}. Numerical and encoded features are standardized with a
scaler fitted only on the training data, which prevents information from
the testing subset from influencing model training.

\subsection{Classification Models}

We compare seven models in this study: a non-learning lazy baseline that
always predicts the majority class, Logistic Regression (LR), and three
Multi-Layer Perceptron (MLP) architectures of increasing capacity, namely
one hidden layer of two neurons, one hidden layer of eight neurons, and
three hidden layers of 64 neurons each. In addition, we include a Random Forest (RF) with
300 trees and a histogram-based gradient boosting classifier (HGB),
because tree ensembles are strong baselines for tabular data.

\subsection{Evaluation Protocol}
\label{sec:protocol}

The cleaned dataset is divided into 80\% training and 20\% testing data
using stratified sampling, and each model is trained and evaluated under
one hundred different random replications, with the same split shared by all
models under each seed. Because accuracy
alone can be misleading under class imbalance \cite{saito2015}, we report
the mean and standard deviation across the runs of the testing accuracy,
balanced accuracy, F1-score for the Air class, the area under the ROC
curve (ROC-AUC), and the area under the precision-recall curve (PR-AUC).
To assess whether the observed differences between models are systematic,
we apply two-sided Wilcoxon signed-rank tests on the paired per-seed
results \cite{demsar2006}.

\section{Explainability and Evaluation Methods}

\subsection{SHAP and LIME}

SHAP assigns a contribution value $\phi_j$ to each feature $j$ for an
individual prediction such that the contributions sum to the difference
between the prediction and a baseline expectation \cite{lundberg2017}. A
positive value moves the output toward the Air class, whereas a negative
value moves it toward the Truck/Ocean class. We compute SHAP values with
the Kernel SHAP algorithm using a background dataset of 25 representative
observations obtained by k-means clustering of the training data, 400
sampled feature coalitions per instance, and 100 test observations per
model and seed, using the first twenty of the one hundred replications to limit
the computational cost. \rev{The explained observations amount to 8.4\% of
the 1,193 observations in each testing split. Every explained
observation requires 10,000 model evaluations under these settings,
i.e., 120 million evaluations for the six models and twenty replications.} Global feature importance is measured by the mean absolute
SHAP value of each feature across the explained observations\rev{, and
thus the global importance vectors and the stability results built on
them derive from this subsample}. LIME
explains an individual prediction by generating perturbed observations
around a selected instance, weighting them with an exponential kernel, and
fitting a weighted linear surrogate \cite{ribeiro2016}. We use the
continuous formulation with 1,000 Gaussian perturbations in the feature space and a ridge-regularized surrogate.

Both methods are implemented from their original formulations in NumPy,
since the analysis requires tight control over backgrounds, sampling, and
replications. To verify the implementation, we compare the Kernel SHAP values
for the logistic regression model with the exact closed-form SHAP values
of a linear model in margin space, and across replications the two agree with a
Pearson correlation of 1.000 and an absolute error below $5\times
10^{-6}$.

\subsection{Stability of Explanations}

If an explanation method is to inform operational decisions, its global
feature ranking should not change materially when the model is
retrained. For each model we therefore compute the Spearman rank
correlation of the global importance vectors between all 190 pairs of
the twenty explanation replications, together with the Jaccard similarity of
the top-5 feature sets. We also compare the seed-averaged
importance vectors across models to quantify how strongly equally
plausible models agree on what matters \cite{breiman2001,krishna2022}.

\subsection{Faithfulness of Explanations}

Faithfulness is evaluated with a feature-deletion test in the spirit of
established attribution benchmarks \cite{samek2017}. For each model, the
top-$k$ features ranked by SHAP and by LIME are replaced by their
training-set means for $k=0,1,\dots,10$, and the mean absolute change of
the predicted Air probability and the fraction of flipped predictions on
the testing data are compared against removal in random order, averaged
over 20 permutations. If the attributions are faithful, removing the
features they rank highest should degrade the predictions considerably
faster than removing random features. \rev{Because 43 of the 47 features
are binary indicators, a training mean is a value that such a feature
never takes in a real shipment. As a complementary scheme we therefore
replace each deleted feature by a value drawn at random from its
training marginal, averaged over five draws, which keeps every indicator
at 0 or 1.}

\section{Results and Discussion}

\subsection{Predictive Performance}

Table~\ref{tab:performance} presents the mean and standard deviation of
the testing metrics across the one hundred runs. The lazy baseline reaches 75.8\%
accuracy by construction with a balanced accuracy of 50\%. The linear and
neural models reach mean testing accuracies between 83.3\% and 85.0\%,
where the two-neuron MLP is the least dependable model, with a testing
standard deviation of 2.3 percentage points that indicates occasional
convergence to poor solutions, while the deep MLP achieves the best
result within the neural family (85.0\%).

\begin{table}[!t]
\caption{Testing performance across one hundred runs (mean $\pm$ standard
deviation, in \%). Best values are in bold.}
\label{tab:performance}
\centering
\scriptsize
\setlength{\tabcolsep}{1.8pt}
\begin{tabular}{lccccc}
\toprule
\textbf{Model} & \textbf{Acc.} & \textbf{Bal. Acc.} & \textbf{F1} &
\textbf{ROC-AUC} & \textbf{PR-AUC} \\
\midrule
Lazy & $75.8 \pm 0.0$ & $50.0 \pm 0.0$ & $86.2 \pm 0.0$ & $50.0 \pm 0.0$ & $75.8 \pm 0.0$ \\
LR & $84.3 \pm 0.8$ & $77.1 \pm 1.4$ & $89.8 \pm 0.6$ & $90.6 \pm 0.8$ & $96.9 \pm 0.3$ \\
MLP (2) & $83.3 \pm 2.3$ & $77.1 \pm 4.3$ & $89.0 \pm 1.6$ & $89.3 \pm 2.2$ & $96.2 \pm 1.0$ \\
MLP (8) & $84.4 \pm 0.9$ & $78.2 \pm 1.6$ & $89.8 \pm 0.6$ & $90.9 \pm 0.8$ & $97.0 \pm 0.4$ \\
MLP (64) & $85.0 \pm 0.9$ & $78.0 \pm 1.8$ & $90.3 \pm 0.6$ & $91.3 \pm 0.8$ & $97.0 \pm 0.4$ \\
RF & $\mathbf{86.8 \pm 0.9}$ & $\mathbf{80.7 \pm 1.4}$ & $\mathbf{91.4 \pm 0.6}$ & $\mathbf{92.8 \pm 0.7}$ & $97.5 \pm 0.3$ \\
HGB & $86.7 \pm 0.9$ & $80.5 \pm 1.5$ & $91.3 \pm 0.6$ & $92.7 \pm 0.7$ & $\mathbf{97.6 \pm 0.3}$ \\
\bottomrule
\end{tabular}
\end{table}

The picture changes once the tree ensembles are included. The Random
Forest attains a mean testing accuracy of 86.8\% and a balanced accuracy
of 80.7\%, and the gradient boosting model follows closely with 86.7\% and
80.5\%. The Wilcoxon signed-rank tests indicate that the Random Forest
outperforms every linear and neural model on all four discriminative
metrics at $p<10^{-16}$, with mean advantages of 1.8 to 3.5 percentage
points in accuracy and 2.6 to 3.7 percentage points in balanced
accuracy (computed from the unrounded
means),
whereas the difference between the two ensembles is small, reaching the
5\% level only marginally on accuracy ($p=0.050$), clearly on balanced
accuracy ($p=0.005$), and not on F1 or ROC-AUC. This means that 
increasing model complexity does not improve generalization on this
dataset, holds within the family of fully connected networks but does not
extend to model families that are better suited to tabular data, in line
with observations that the choice among good models is rarely
neutral \cite{rudin2019}.

\subsection{Global Explanations Across Models}

\begin{figure}[!t]
\centering
\includegraphics[width=\columnwidth]{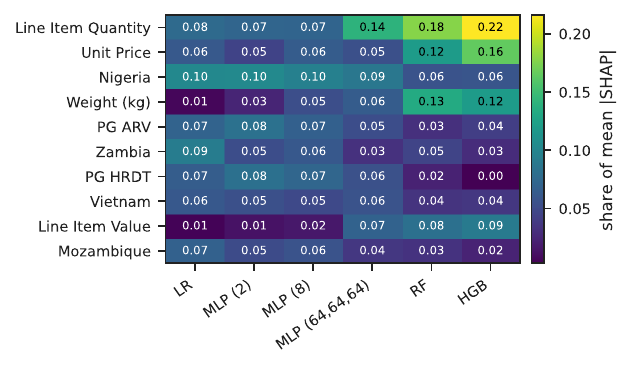}
\caption{Share of global SHAP attribution per feature (seed-averaged) for
the six trained models.}
\label{fig:global}
\end{figure}

Figure~\ref{fig:global} summarizes the seed-averaged global SHAP
importance, expressed as each feature's share of the total attribution
mass, for the twelve features with the largest overall share. With
\textit{Pack Price} removed, the models divide into two groups.
The linear model and the two smaller MLPs allocate much of their
attribution to the destination indicators, led by Nigeria and Zambia, and
to the product-group indicators. The deep MLP and the tree ensembles
instead concentrate on the continuous shipment attributes, with
\textit{Line Item Quantity}, \textit{Unit Price}, \textit{Weight
(Kilograms)}, and \textit{Line Item Value} ranked highly. It is
interesting to note that the models emphasizing the operational shipment
attributes are also the more accurate ones, so accuracy and the
substantive content of the explanations move together on this dataset.

\subsection{Geographic Distribution of Destination Effects}

\begin{figure*}[!t]
\centering
\includegraphics[width=0.9\textwidth]{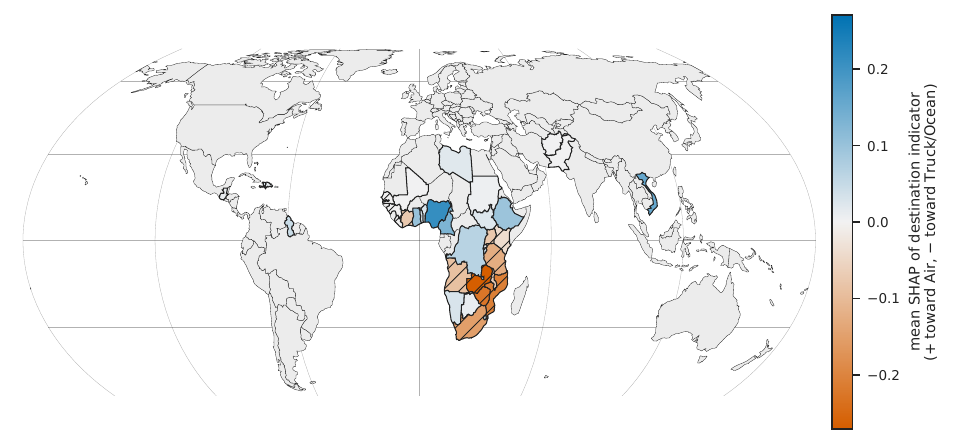}
\caption{Mean SHAP contribution of each destination-country indicator for
shipments destined to that country (Random Forest, five replications), on the
equal-area Equal Earth projection. Hatched countries push predictions
toward Truck/Ocean; countries in gray do not appear in the dataset.}
\label{fig:worldmap}
\end{figure*}

Because every destination country is encoded as its own indicator, the
geographic pattern of the destination effects can be examined for all 38
countries in the dataset. For each country we average the SHAP value of
its own indicator over the test shipments destined there, using the
Random Forest as the most accurate model across five replications. \rev{The map explains every test shipment of a seed, which costs about
twelve times as much per seed as the 100-observation subsample of the
stability analysis. We therefore limit it to five replications.}
\rev{As a check on this choice, we recomputed the
estimates with all twenty explanations. The ten estimates quoted
below moved by at most 0.01, none changed sign, and their
standard deviation across individual replications stayed below 0.02.}
Figure~\ref{fig:worldmap} displays the result on the equal-area Equal
Earth projection \cite{savric2019}, which avoids the severe area
distortion of the Mercator projection.

The map reveals a coherent regional structure. The strongest effects
toward Truck/Ocean form a contiguous southern and eastern African
cluster, led by Zambia ($-0.27$\rev{, $n=435$}), Zimbabwe ($-0.22$\rev{,
$n=309$}), Mozambique ($-0.21$\rev{, $n=360$}), South Africa
($-0.15$\rev{, $n=282$}), and Tanzania ($-0.13$\rev{, $n=398$}), while the
strongest effects toward Air appear for Nigeria ($+0.21$\rev{, $n=748$}),
Vietnam ($+0.16$\rev{, $n=445$}), Cameroon ($+0.13$\rev{, $n=55$}), Ghana
($+0.10$\rev{, $n=42$}), and Ethiopia ($+0.10$\rev{, $n=164$})\rev{, where
$n$ is the number of shipments to that country in the cleaned data and
about one fifth of them fall in each testing split}.
Simple port access does not explain this pattern: landlocked and coastal
destinations do not differ significantly in their mean contributions
(Mann-Whitney $p=0.45$), since landlocked Zambia and Zimbabwe push
strongly toward surface transport while landlocked Ethiopia and Burundi
lean toward air. A more plausible reading is that the surface-leaning
cluster coincides with the established road corridors that connect the
ports of Durban, Beira, and Dar es Salaam to the southern African
interior, whereas destinations such as Nigeria may face port congestion,
longer or less reliable inland corridors, and security concerns that make
air freight attractive for time-sensitive medical cargo, consistent with
the view that trade logistics performance depends on corridor
infrastructure and border efficiency and not on coastline alone
\cite{arvis2018}. The SHAP values themselves establish no causal claim,
and countries with very few shipments receive estimates close to zero.
\rev{The shipment counts are highly uneven, since the eleven largest
destinations each have more than 280 shipments and account for 83\% of
the cleaned records, whereas fourteen of the 38 countries have fewer
than 20. The estimates for Cameroon and Ghana
rest on roughly ten test shipments per seed and should be read as
indicative only. The same caution applies to Burundi ($n=56$) in the
port-access comparison.}

\subsection{Stability Across replications and Models}

\begin{figure*}[!t]
\centering
\includegraphics[width=\textwidth]{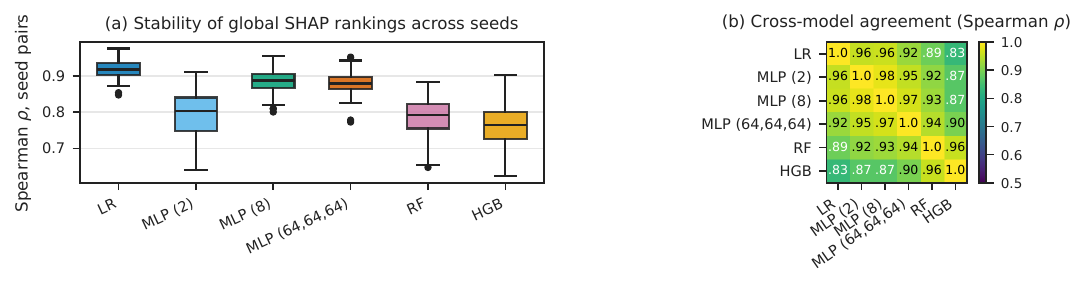}
\caption{Stability of global SHAP explanations: (a) rank correlation
between seed pairs for each model and (b) rank correlation between the averaged rankings.}
\label{fig:stability}
\end{figure*}

Figure~\ref{fig:stability}(a) shows the distribution of Spearman rank
correlations between the global SHAP rankings of all 190 seed pairs, and
the two stability measures capture different aspects of the rankings.
Measured over all 47 features, the linear model is the most stable
($\rho=0.917$\rev{, with an interquartile range of 0.903 to 0.934 across
the 190 seed pairs}), while the tree ensembles average 0.76 to 0.79
\rev{(interquartile ranges of 0.727 to 0.801 for HGB and 0.755 to 0.823
for RF)} because the
many near-zero features in their long importance tail rank noisily.
Measured on the decision-relevant top of the ranking, the order reverses:
the top-5 sets of HGB and RF are the most stable across retraining
(Jaccard 0.80 and 0.76), whereas the baseline MLP retains only 0.38.
\rev{The corresponding standard deviations across the 190 seed pairs are
0.17, 0.13, and 0.19.}
Therefore, the leading features of the tree ensembles can be reported
with more confidence than those of a single trained MLP.

Figure~\ref{fig:stability}(b) reports the agreement between the
seed-averaged rankings of different models. The full-ranking correlations
appear high ($\rho$ between 0.83 and 0.98) because all models agree that
most of the 47 features barely matter. The agreement on the features that
do matter is much weaker: the mean top-5 Jaccard similarity across model
pairs is only 0.44\rev{, with the fifteen individual pairs ranging from
0.11 to 1.00}, and the lowest full-ranking agreement occurs between
LR and HGB, the pair whose attribution mass is divided most differently
between geography and shipment attributes. In other words, models whose
testing accuracies differ by less than four percentage points still
disagree substantially on which features matter most. This finding
provides concrete, domain-specific evidence for the disagreement problem
described by Krishna et al.~\cite{krishna2022} and for the Rashomon
effect \cite{breiman2001}: the data admit several similarly
accurate descriptions that tell different operational stories.

\subsection{Faithfulness of the Attributions}

\begin{figure}[!t]
\centering
\includegraphics[width=\columnwidth]{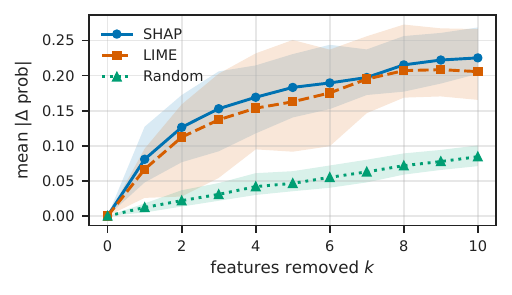}
\caption{Faithfulness deletion test. Mean absolute change of the predicted
Air probability when the top-$k$ features ranked by SHAP, LIME, or a
random ordering are replaced by their training means. \rev{Shaded
bands show the range across the six models.}}
\label{fig:faithfulness}
\end{figure}

Figure~\ref{fig:faithfulness} presents the deletion curves averaged over
the six models. For every model, removing the features ranked most
important by SHAP or LIME degrades the predictions much faster than
removing random features. Deleting the top five SHAP-ranked features
changes the predicted probabilities by 18.3 percentage points on average
and flips 22.5\% of the predictions, against 4.7 points and 5.2\% for the
random baseline, which corresponds to roughly a fourfold separation, and
the LIME rankings perform comparably (16.3 points, 17.8\%). \rev{The
shaded bands in Figure~\ref{fig:faithfulness} give the range across the
six models}\rev{. At $k=5$ the change spans 14.0 to 24.4 points for SHAP and
9.2 to 23.7 points for LIME, against 3.5 to 7.2 points for random
removal. The least affected model under either method thus still changes
more than the most affected model under random removal.}
\rev{The separation persists under the marginal replacement scheme.
Deleting the top five SHAP-ranked features then changes the predicted
probabilities by 19.3 points and the top five LIME-ranked
features by 17.9 points, against 5.9 points for random
removal, with standard deviations of 0.8, 1.2, and
0.8 points across the twenty replications. The individual random
permutations vary more widely, with a standard deviation of 3.5
points. Even so, the SHAP ranking exceeds every one of the twenty
permutations in all 120 combinations of model and seed.
The faithfulness result therefore holds when every perturbed indicator
stays at 0 or 1.} The wide gap
confirms that both methods concentrate on features the models actually
use, so the disagreement observed above reflects real differences between
the models and not artifacts of the explanation procedure.

\subsection{Local Explanations for a Selected Shipment}

\begin{figure}[!t]
\centering
\includegraphics[width=\columnwidth]{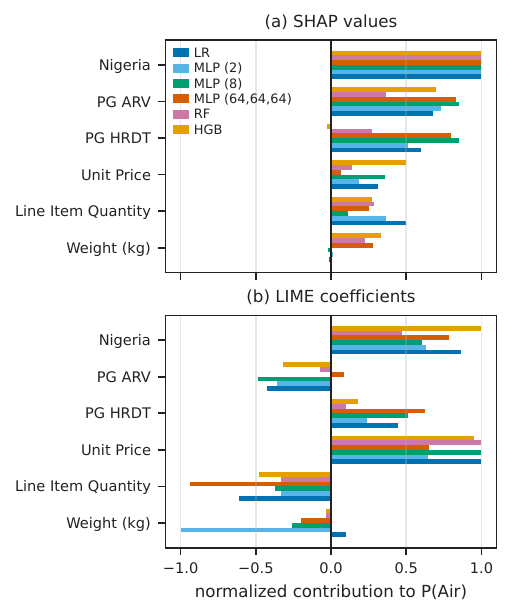}
\caption{Local explanations of the same selected shipment across models:
(a) SHAP values and (b) LIME coefficients, each normalized by the largest
absolute contribution of the corresponding model.}
\label{fig:local}
\end{figure}

To connect the aggregate analysis with the practitioner's view of a single
decision, we explain one selected test shipment across all six models. The
shipment is an Air delivery of rapid test kits (product group HRDT) to
Nigeria with a line-item quantity of 7,301, a unit price of 1.35, a
weight of 2,093 kilograms, and a line-item value of 197,127. Every model
predicts the Air class with a probability of essentially 1.00.
Figure~\ref{fig:local} compares the SHAP values and the LIME coefficients
for the six features with the largest average contribution.

For this shipment the six models agree on the principal supporting
factor, since the Nigeria indicator carries the largest positive SHAP
contribution in every model, followed by the product-group indicators,
which is consistent with the strong Air tendency of Nigerian
destinations in Figure~\ref{fig:worldmap}. The comparison between the
two methods reveals a subtlety. For most models, SHAP assigns
\textit{Line Item Quantity} a positive contribution while LIME assigns
it a negative coefficient. The two statements are not contradictory,
because SHAP measures the contribution of the observed quantity relative
to a background population, whereas the LIME coefficient measures the
local slope. However, a practitioner who reads the two outputs side by
side without this distinction could easily draw opposite conclusions
from the same shipment, which reinforces the need to accompany post-hoc
explanations with their precise semantics \cite{salih2025,krishna2022}.

\subsection{Practical Implications and Limitations}

\rev{Four} practical implications follow from the results. First, model
selection for tabular freight data should include tree ensembles before
any conclusion about the value of model complexity is drawn, since the
ensembles here are simultaneously the most accurate models and the models
whose leading features remain most stable across retraining. Second,
explanation stability is inexpensive to measure and should be reported
alongside accuracy, because unstable rankings, as observed for the small
MLPs, undermine any operational insight built on a single trained model.
Third, the geographic structure of the destination effects suggests that
corridor-level variables, such as distance to the serving port, corridor
condition, and route security, would be informative
additions to shipment-level data for mode-choice modeling
\cite{arvis2018}. \rev{Fourth, when equally accurate candidate models
disagree on feature importance, a logistics team should act on the
features that lead in every candidate model, such as \textit{Line Item
Quantity} and the Nigeria indicator in Figure~\ref{fig:global}. A
feature that leads for one model only is better handled as a hypothesis
to be checked against corridor knowledge or additional data before it
informs a routing or procurement decision.} These interpretations describe how the trained models
use the available features and should not be read as causal statements
\cite{schwalbe2024,xu2024}.

This study has several limitations. The dataset covers a single
humanitarian health supply chain, so the specific feature rankings may
not transfer to commercial freight settings. \rev{The same caution
applies to the destination effects, which need validation on a second
freight dataset before they are generalized to other cargo or corridors. The binary target
merges Air with Air Charter and Truck with Ocean, and thus the
explanations cannot separate the individual modes, of which Ocean
accounts for only 282 of the 5,964 cleaned records.} Important transportation
variables such as distance and delivery time are not available, while
\textit{Freight Cost} and \textit{Pack Price} are intentionally excluded.
The stability analysis considers retraining variability under a fixed
preprocessing pipeline, and the deletion test replaces feature values
with training means, which is one of several possible reference
distributions for measuring faithfulness. \rev{Mean replacement is
particularly awkward for the 43 binary indicators, i.e., 91\% of the
features, whose mean is a fraction that no real shipment takes. The marginal-replacement check removes this specific
problem, but it still ignores the dependence between features and can,
for example, activate two country indicators at once. Conditional
replacement and counterfactual perturbation would close this remaining
gap. Finally, larger explanation samples and alternative background
sets are robustness checks that we leave for future work.}

\section{Conclusions}

Machine learning is increasingly embedded in freight transportation
decisions, but the explanations that accompany its predictions usually
reach practitioners without any evidence of their reliability. Whether
these explanations can be trusted is therefore as important as whether
the predictions are accurate. This study investigated explainable
machine learning for binary freight transportation mode classification
on the PEPFAR SCMS Delivery History dataset, treating the explanations
themselves as objects of evaluation. Seven models were compared with imbalance-aware metrics and paired significance
tests, and the SHAP and LIME explanations were assessed for
stability, faithfulness, and geographic structure.
The tree ensembles significantly outperformed the linear and neural
models, reaching 86.8\% mean testing accuracy against 83.3--85.0\%, which
shows that the earlier conclusion about the limited value of model
complexity was specific to fully connected networks. The leading features
of the ensembles were also the most stable across retraining, whereas the
small MLP combined the weakest performance with the least dependable
rankings. Models with similar accuracy still disagreed substantially
about which features matter most, with a mean top-5 overlap of only 0.44
between model pairs, while the deletion test confirmed that both SHAP and
LIME identify features the models rely on, degrading predictions roughly
four times faster than random feature removal. Encoding every
destination country revealed a coherent geographic pattern, with the
southern African corridor destinations pushing predictions toward
surface transport and destinations such as Nigeria and Vietnam pushing
them toward air, a pattern that port access alone does not explain.
Overall, our findings indicate that explanations in freight
transportation analytics should be treated as measured quantities with
their own quality criteria.

\section*{Acknowledgment}

The authors acknowledge the support of the the
Inbound Summer Research Program at King Fahd University of Petroleum and Minerals (KFUPM), Saudi Arabia.

\bibliographystyle{IEEEtran}
\bibliography{references}

\end{document}